\documentclass[runningheads]{llncs}
\usepackage[T1]{fontenc}
\usepackage{enumitem}
\usepackage{setspace}
\usepackage{multirow}
\usepackage{pgfplots}
\usepackage{graphicx}
\usepackage{amsmath}
\usepackage{amssymb}
\usepackage{listings}
\definecolor{codegreen}{rgb}{0,0.6,0}
\definecolor{codegray}{rgb}{0.5,0.5,0.5}
\definecolor{codepurple}{rgb}{0.58,0,0.82}
\definecolor{backcolour}{rgb}{0.95,0.95,0.92}
\lstdefinestyle{mystyle}{
    commentstyle=\color{codegreen},
    keywordstyle=\color{magenta},
    numberstyle=\tiny\color{codegray},
    stringstyle=\color{codepurple},
    basicstyle=\ttfamily\footnotesize,
    breakatwhitespace=false,         
    breaklines=true,                 
    captionpos=b,                    
    keepspaces=true,                 
    numbers=left,                    
    numbersep=5pt,                  
    showspaces=false,                
    showstringspaces=false,
    showtabs=false,                  
    tabsize=2, 
    morekeywords={parallel_for}, 
    literate={parallel_for}{{\textcolor{magenta}{parallel\_for}}}9
}

\begin{document}
\title{A High-Level Compiler Integration Approach for Deep Learning Accelerators Supporting Abstraction and Optimization}
%
%
\author{Samira Ahmadifarsani\inst{1}\orcidID{0009-0003-4517-168X} \and \\
Daniel Mueller-Gritschneder\inst{2}\orcidID{0000-0003-0903-631X} \and \\
Ulf Schlichtmann\inst{1}\orcidID{0000-0003-4431-7619}}
\authorrunning{S. Ahmadifarsani et al.}
%
\institute{Technical University of Munich, Arcisstr. 21, 80333 Munich, Germany
\email{\{samira.ahmadifarsani, ulf.schlichtmann\}@tum.de}\and
TU Wien, 1040 Vienna, Austria\\
\email{daniel.mueller-gritschneder@tuwien.ac.at}}
\maketitle              
\begin{abstract}
The growing adoption of domain-specific architectures in edge computing platforms for deep learning has highlighted the efficiency of hardware accelerators. However, integrating custom accelerators into modern machine learning (ML) compilers remains a complex challenge due to the need for significant modifications in compilation layers and specialized scheduling techniques. Existing frameworks offer partial solutions and require users to navigate intricate compiler internals.

In this paper, we introduce a TVM-based compilation integration approach that targets GEMM-based deep learning accelerators. Our approach abstracts the complexities of compiler integration, enabling seamless integration of accelerators without requiring in-depth knowledge of the underlying compiler. Furthermore, we extend and incorporate design space exploration tools, specifically CoSA, to automate efficient tensor scheduling, accounting for factors such as uneven mapping and double buffering. Our framework is benchmarked on the Gemmini accelerator, demonstrating performance comparable to its specialized manually implemented toolchain.
\end{abstract}
\section{Introduction}
Domain-specific architectures have proven effective in edge computing for deep learning, offering enhanced power-performance through customized compute engines and memory hierarchies. Spatial architectures, like Google's TPU and Gemmini~\cite{Gemmini}, excel at tensor-based tasks using systolic arrays optimized for GEMM operations. To address limited on-chip memory, accelerators often replace caches with software-managed scratchpads, relying on explicit DMA-based transfers. They are typically paired with general-purpose processors that manage unsupported tasks and coordinate control via custom instructions.

Despite advancements in hardware design tools, deploying deep learning models on custom accelerators is hindered by the need for complex software toolchains. ML compilers like TVM~\cite{TVM} and TinyIREE~\cite{tinyiree} offer robust support for CPUs and GPUs, but integrating new accelerators demands deep compiler modifications. A key challenge is tensor scheduling, which controls computation and data movement. While some compilers use search-based strategies (e.g., Ansor~\cite{ansor}), these fall back to templates for accelerators, requiring manual tuning for each intrinsic.



This paper presents a compiler integration approach for GEMM-based hardware accelerators that simplifies the integration process by abstracting low-level compilation complexities. Our key contributions are:

\begin{itemize}
    \item \vspace{-5pt}We utilize established design space exploration (DSE) tools and their unified input representations to efficiently schedule tensor operations. Specifically, we extend CoSA~\cite{CoSA} to initialize scheduling parameters, accounting for factors like uneven mapping and double buffering. 
    \item We propose a compact description for GEMM-based hardware accelerators, enabling users to define the functionality and programming interface for convolution and dense operators.
    \item  We present an automated flow, comprising a Frontend Configurator and a Backend Configurator, that generates a TVM-based compiler backend with minimal manual effort, unlike existing methods that branch out to custom backends.  
\end{itemize} 

We benchmark our work with a case study on the Gemmini hardware accelerator, achieving performance comparable to its manually implemented C-function-based toolchain while reducing the manual effort by 80 percent, simply looking at lines of code. It is also important to note that the complexity of compiler integration is significantly reduced.
\section{Related Work}
TVM’s Bring Your Own Code (BYOC)~\cite{BYOC} enables integration of custom accelerators or kernel libraries at both graph and operator levels. UMA~\cite{UMAhow} builds on BYOC by introducing structured configuration files and Python APIs to expose backend interfaces. While UMA works well as an interface for simple accelerator models, it lacks support for essential features like quantization, scheduling, tiling, and fine-grained instruction mapping, making it inadequate for real-world accelerators~\cite{mypaper}. Users are still required to navigate the complexities of TVM’s integration pipeline.

Similarly, MATCH~\cite{MATCH} offers a customizable, model-driven abstraction built on Relay IR, the high-level intermediate representation (IR) of TVM, and generates runtime modules through a C-code backend based on MAKO templates. However, it requires users to define network transformations to rewrite the computational graph based on hardware-specific characteristics, demanding familiarity with compiler internals.

\section{Proposed Approach}
\begin{figure}[tb]
\centering
\includegraphics[width=\textwidth, height=6cm, keepaspectratio]{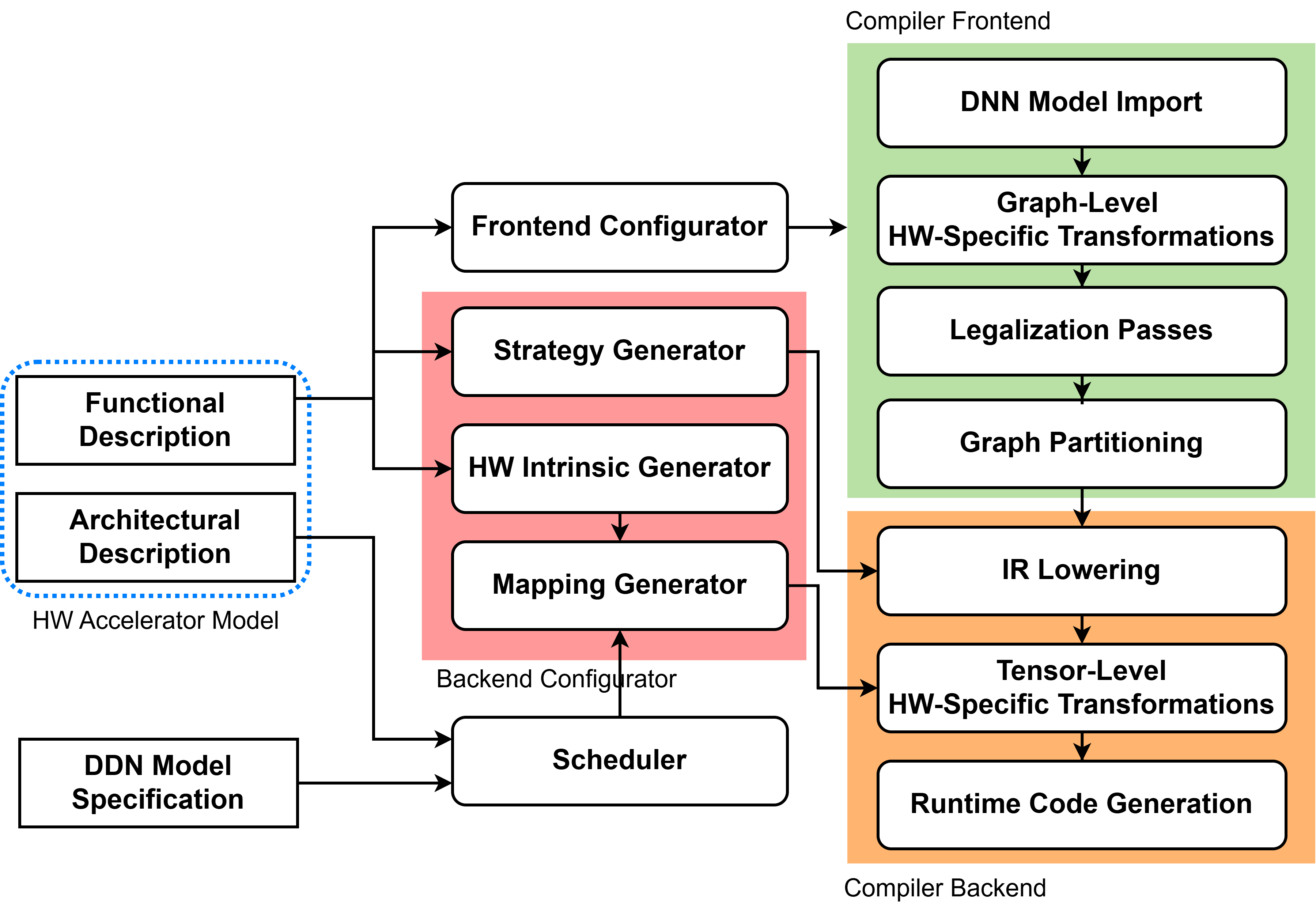}
\caption{Proposed compiler integration framework.} \label{fig3}
\end{figure}

This paper presents a specialized compilation approach for custom hardware accelerators, as shown in Fig.~\ref{fig3}, which includes two main stages: the compiler and its configurators. The system requires two user inputs: a hardware model and DNN specifications. Our method extends UMA to address key integration challenges in TVM. In the following sections, we provide an overview of the scheduling process, hardware model and configurators, which play a crucial role in the overall system.

\vspace{-10pt}
\subsection{Extending CoSA Scheduler for Tensor Scheduling}
For GEMM-based hardware accelerators, specialized architectures and low-level programming interfaces limit execution to a predefined set of dataflows. Additionally, valid mappings must adhere to both compute and memory constraints. This work extends the CoSA scheduler~\cite{CoSA} to efficiently generate high-performance schedules from our Accelerator Description. CoSA is a constrained optimization method that formulates DNN scheduling as a Mixed-Integer Programming (MIP) problem for spatial hardware accelerators. CoSA takes as input the workload’s shape and parameters, hardware specifications (e.g., compute units, memory, network topology), and constraints (e.g., memory-level skipping). These are defined using a user-friendly YAML configuration format.

\begin{figure}[tb]
\centering
\includegraphics[width=\textwidth, height=3cm, keepaspectratio]{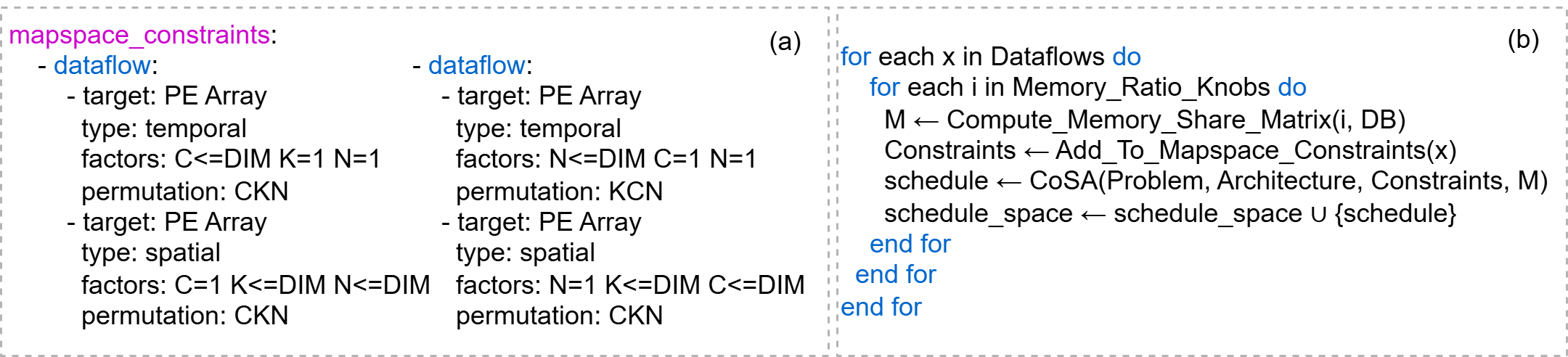}
\caption{a) Output (left) and weight (right) stationary dataflow examples and loop factor limitations as a set of hardware constraints. b) pseudo code for generating the final schedule space using extended CoSA.} \label{datflow-fig}
\end{figure}
By default, CoSA assumes a fully flexible hardware organization, exploring all mappings generated by loop transformations like tiling, permutation, and spatial mapping. However, many of these mappings are not directly executable on our targeted GEMM-based hardware accelerators, which are already statically and spatially laid out. Moreover, we must impose limits on loop factors to ensure compatibility with the accelerator’s instruction set. For example, the compute instructions of the Gemmini accelerator perform GEMM\setcounter{footnote}{0}\footnote{Consider a GEMM operation where $In \in \mathbb{R}^{N \times C}$, $W \in \mathbb{R}^{C \times K}$, and $O \in \mathbb{R}^{N \times K}$.} operations with N, C, and K smaller than the dimension of the PE array ($\text{DIM}\times\text{DIM}$). 

To generate fitting schedules, we first model accelerator-specific dataflows and the constraints imposed by the accelerator's instruction set as part of CoSA’s constraint inputs, as shown in Fig.~\ref{datflow-fig} (a). Users can define architecture constraints, along with fixed spatial and temporal settings, within these constraints as part of the accelerator’s architectural description. Secondly, we extend the CoSA MIP model to incorporate these constraints such as below\footnote{Due to space limitations, we omit background details; for relevant context that aids in understanding this equation, please refer to the CoSA paper~\cite{CoSA}.}, ensuring that at the PE array level (denoted as \textit{I}), the spatial and temporal loop bounds for each dimension (N, C, K, denoted as \textit{J}) do not exceed DIM. X represents the schedule space as a binary 4D matrix, with dimensions representing (1) layer dimension variables ($j$), (2) prime factors of loop bounds ($n$), (3) memory and permutation levels ($i$), and (4) spatial or temporal mapping ($k$).
\begin{equation}
    \sum_{n,k}\log(prime\_factor_{J,n})X_{J,n,I,k}\leq \log(\text{DIM})
\end{equation}

In the final scheduling process, an optimized search space is generated by running the extended CoSA across all valid combinations of tuning parameters, including accelerator-supported dataflows, uneven mapping strategies, and double buffering, as shown in Fig.~\ref{datflow-fig} (b). In the original CoSA, memory constraints are defined using a fixed array that specifies the share of each memory level allocated to each operand. We leverage this array to explore different memory share configurations for input, weight, and output tensors at each level, enabling support for uneven mapping. When double buffering is supported, we halve the maximum available memory for each operand to ensure they fit within half of the on-chip memories. The refined CoSA outputs are then passed to the mapping generator, which translates them into Tensor IR (TIR) transformations. Finally, the generated schedules, including intrinsic calls, are evaluated on the hardware to determine the most efficient configuration based on real execution profiling.

\subsection{GEMM-based Hardware Accelerator Descriptions}\label{secdes} \vspace{-5pt}
Each compilation stage requires hardware-specific inputs. IR lowering relies on operator implementations detailing how operators execute on the accelerator. Loop scheduling depends on architectural features, and code generation identifies interface functions for accelerator invocation. To supply these, users define a hardware accelerator model comprising functional and architectural descriptions.

\begin{figure}[tb]
\centering
\includegraphics[width=\textwidth, keepaspectratio]{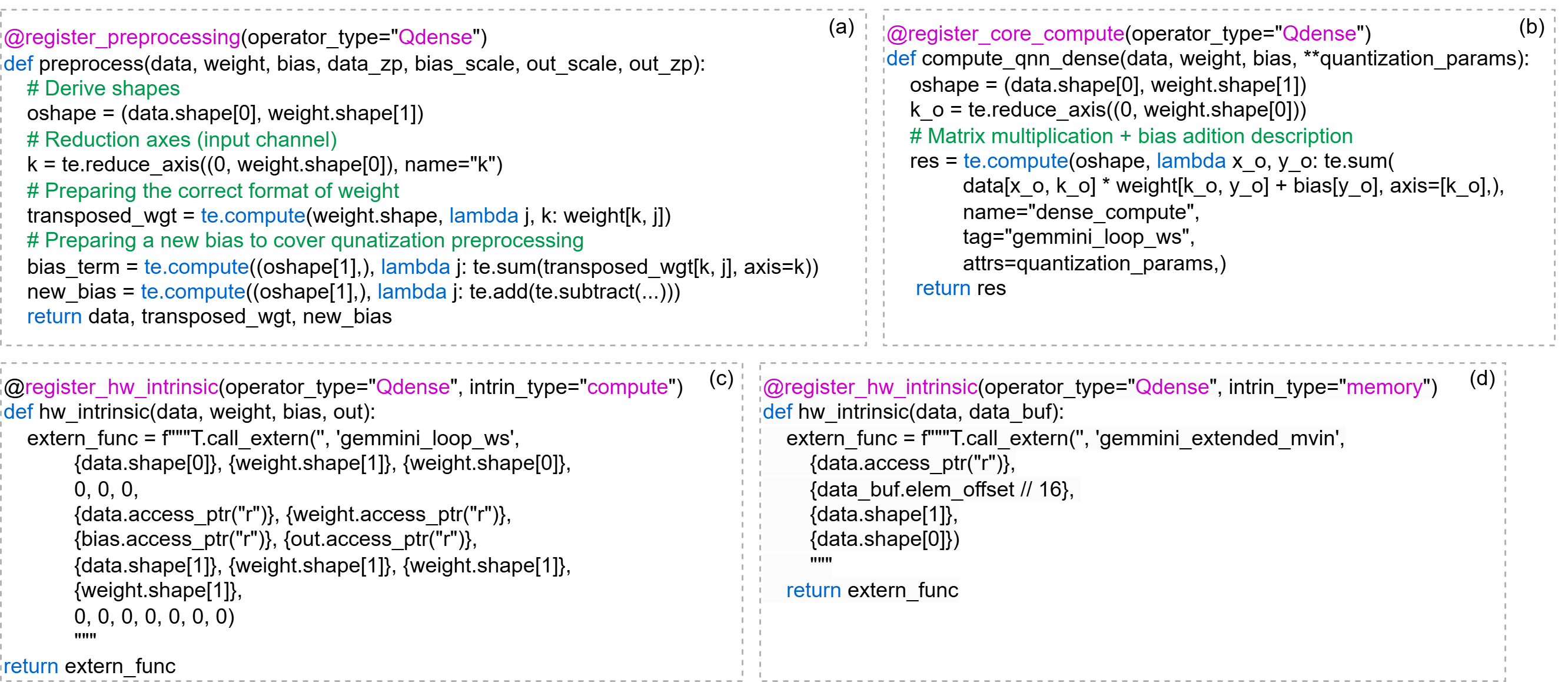}
\caption{Examples for registering a) the preprocessing, b) core computations of a dense layer, c) matrix multiplication, and d) memory load intrinsics in Gemmini.} \label{fig5}
\end{figure}

The functional description specifies supported operators through two key components: preprocessing and compute functions. While GEMM-based accelerators support dense and convolution layers, effective deployment often requires transformations like transposition, flattening, or im2col. Operator implementations use Tensor Expression (TE), a Python DSL for computation. Fig.~\ref{fig5} (a) and (b) show a quantized dense operator example. To ease development, we provide Python APIs for registering these functions via decorators like @register\_core\_compute and @register\_preprocessing. Constant-related preprocessing is folded at compile-time; others run on the host CPU. Each compute function is linked to an accelerator interface via a user-defined tag and consumed by the strategy generator.

The second part defines interface functions for offloading operations. Accelerators vary in control granularity—from fine-grained instruction sets to high-level operations limited by on-chip memory. Users register these intrinsics using @register\_hw\_intrinsic, categorized into compute, memory, and configuration intrinsics. Fig.~\ref{fig5} (c) and (d) illustrate the registration of matrix multiplication and memory intrinsics of Gemmini.

The architectural description provides the necessary information for scheduling and follows the same format as the input required by CoSA. This includes YAML template files that specify (a) the hardware organization, detailing the topology of compute and storage units, and (b) hardware constraints, which define limitations on the set of valid mappings for the hardware.

\subsection{Accelerator Integration Steps} \label{config_sec}
With a compilation pipeline tailored for hardware accelerators in place, the configurators automate essential tasks, leveraging hardware accelerator descriptions to streamline their integration into the compiler.
\vspace{-12pt}
\subsubsection{Frontend Configurator} TVM’s import module typically parses a quantized operator as a sequence of operations (e.g., for a TFLite dense op: QNN dense, bias add, requantize, clip). However, since TVM lowers Relay to TIR at the operator level, it cannot generate a unified function for such multi-op sequences. Solving this typically requires custom Relay ops and hardware-specific legalization passes~\cite{Gemmini-BYOC}. To simplify, we introduce generalized Relay operators (e.g., dense, convolution) that encapsulate full sequences. Once marked as supported, a legalization pass rewrites the sequence into a single operator, enabling TIR lowering. Hardware-specific implementations are incorporated later in Strategy Generator via user-defined descriptions. 

With these modifications in the compilation process, the frontend now requires minimal user intervention. The frontend configurator sets up the graph partitioning and legalization passes using predefined supported operators, derived from the functional description of the hardware accelerator. This ensures that the accelerator-specific Relay operators are correctly identified and transformed before graph partitioning.
\vspace{-12pt}
\subsubsection{Strategy Generator} 
In the backend, IR lowering requires a well-defined strategy that consists of a tensor computation description and its scheduling.
The strategy generator creates the strategy by binding the user-defined computation function and a default schedule to the corresponding operator and ensuring seamless integration with the compilation pipeline. Unlike traditional TVM lowering, where TE schedules are explicitly defined, UMA bypasses TE scheduling. While this could be seen as a limitation, we turn it into an opportunity by handling scheduling at the TIR level via the Mapping Generator. This approach not only provides greater flexibility but also makes debugging significantly easier in TVM. 
\vspace{-12pt}
\subsubsection{Hardware Intrinsic Generator} TVM provides a tensorization interface that allows users to manually invoke intrinsics when implementing software. However, defining hardware intrinsics in a way that the compiler can interpret requires additional effort. Specifically, users must register a tensor intrinsic that includes both a computation description and its corresponding implementation. The system utilizes this description to identify relevant computation regions, while the implementation maps the computation to specialized hardware instructions. Instead of requiring manual registration, the hardware intrinsic generator leverages the user-defined functional description in the accelerator model to automatically generate the necessary tensor intrinsics for the compiler.
\vspace{-12pt}
\subsubsection{Mapping Generator} The next step in the backend is scheduling, which applies TIR-level hardware-specific transformations. Scheduling decisions, including multi-level tiling and reordering, are generated using the extended CoSA scheduler. CoSA produces a YAML file that specifies the tile factors and the ordering of tensor dimensions for each memory level. Based on this output, the mapping generator applies loop transformations using TIR schedule primitives. At this stage, the insertion of hardware intrinsics must also be performed. After applying the loop-level schedules, the mapping generator utilizes TVM’s tensorization feature to rewrite TIR stages with hardware intrinsics. This is done using TIR intrinsics that were previously generated by the hardware intrinsic generator.
{\setstretch{0.9}\section{Evaluation}
To build a TVM-based backend for Gemmini using our method, users need only provide a functional description in Python and an architectural configuration in YAML. As shown in Table~\ref{tab2}, this reduces manual effort by 80\% in terms of Lines of Code (LoC) for enabling IR lowering and scheduling, compared to traditional integration, and removes the need for deep TVM compiler knowledge.}

We evaluate performance by running single dense layers of varying sizes and the full ToyCar network from the MLPerf benchmark on a cycle-accurate Gemmini simulator via Verilator. Inference latency is compared against two baselines: (1) a manually integrated backend using UMA/BYOC, and (2) our backend without scheduling, relying on Gemmini’s optimized C functions. These functions support large GEMM tiling and efficient loop instruction invocation. We evaluate the weight-stationary kernel configuration, the most performant setup in Gemmini’s default toolchain, against our scheduler’s generated mappings.
\begin{table}[tb]
\centering
\caption{Estimation of LoC for manually enabling lowering and scheduling in Gemmini backend for 2D convolution and dense operators.}\label{tab2}
\resizebox{\textwidth}{!}{%
\begin{tabular}{|cc|c|c|l|}
\hline
\multicolumn{2}{|c|}{\textbf{Lowering (Manual)}}         & \textbf{Scheduling (Manual)} & \textbf{Proposed}                                    & \multirow{2}{*}{\textbf{Reduction}} \\ \cline{1-4}
\multicolumn{1}{|c|}{Relay IR (C++)} & Relay IR (Python) & TE/TIR (Python, TVMScript)   & \multicolumn{1}{l|}{Functional Description (Python)} &                                     \\ \hline
\multicolumn{1}{|c|}{$\sim$230 LoC}  & $\sim$398 LoC     & $\sim$425 LoC                & $\sim$208 LoC                                        & \multicolumn{1}{c|}{$\sim$80\%}     \\ \hline
\end{tabular}%
}
\end{table}

As shown in Table~\ref{tab3}, our toolchain with automated accelerator integration achieves performance comparable to Gemmini’s manually optimized C-function-based toolchain. In contrast, the naive UMA/BYOC-based backend shows significant performance degradation across all test cases. This drop stems from inefficient handling of preprocessing operations, such as matrix transposition and quantization, which, without proper constant folding, impose substantial overhead. TVM typically disables constant folding for matched operators after graph partitioning, and re-enabling it is non-trivial. We addressed this by extending UMA’s Lower module to extract and propagate constant parameters correctly, and by modifying TVM’s low-level structure checks to apply constant folding.
\begin{table}[tb]
\scriptsize
\centering
\caption{Deployment results.}\label{tab3}
\begin{tabular}{c|ccc|}
\cline{2-4}
\textbf{}                                                                                        & \multicolumn{3}{c|}{\textbf{Latency (Cycles)}}                                                                                            \\ \hline
\multicolumn{1}{|c|}{\textbf{\begin{tabular}[c]{@{}c@{}}Single Layer \\ (N, K, C)\end{tabular}}} & \multicolumn{1}{c|}{\textbf{$\quad$Proposed$\quad$}} & \multicolumn{1}{c|}{\textbf{C-based Toolchain}} & \multicolumn{1}{l|}{\textbf{BYOC/UMA Backend}} \\ \cline{2-4} 
\multicolumn{1}{|c|}{(64, 64, 64)}                                                               & \multicolumn{1}{c|}{69,995}             & \multicolumn{1}{c|}{69,994}                      & \multicolumn{1}{c|}{160,163}                                                \\
\multicolumn{1}{|c|}{(128, 128, 128)}                                                            & \multicolumn{1}{c|}{279,206}            & \multicolumn{1}{c|}{280,598}                     & \multicolumn{1}{c|}{843,481}                                                \\
\multicolumn{1}{|c|}{(256, 256, 256)}                                                            & \multicolumn{1}{c|}{1,138,769}           & \multicolumn{1}{c|}{1,139,145}                    & \multicolumn{1}{c|}{4,261,116}                                                \\
\multicolumn{1}{|c|}{(512, 512, 512)}                                                            & \multicolumn{1}{c|}{4,877,499}           & \multicolumn{1}{c|}{4,892,657}                           & \multicolumn{1}{c|}{21,508,629}                                                \\ \hline
\multicolumn{1}{|c|}{ToyCar}                                                                     & \multicolumn{1}{c|}{50,064}             & \multicolumn{1}{c|}{51,034}                      & \multicolumn{1}{c|}{10,136,186}\\ \hline
\end{tabular}%
\end{table}
\vspace{-10pt}
\section{Conclusion} \vspace{-5pt}
We introduced an automated compiler integration approach for GEMM-based accelerators, enabling seamless integration into TVM without major changes to its compilation layers. By extending UMA for quantized models and incorporating CoSA for hardware-aware scheduling, our method efficiently maps tensor operations while respecting deployment constraints. Benchmarks on the Gemmini accelerator show performance comparable to its specialized toolchain, with added flexibility and automation. Future work includes broader accelerator support and enhanced scheduling exploration.
%
%
%
\bibliographystyle{splncs04}
\bibliography{references}
\end{document}